# Multispectral Object Detection with Deep Learning

Md Osman Gani[1], Somenath Kuiry[2], Alaka Das[2], Mita Nasipuri[1], Nibaran Das[1],

[1] Department of CSE, Jadavpur University, Kolkata-700032, INDIA
Mdosmangoni28@gmail.com, mitanasipuri@yahoo.com,
nibaran.das@jadavpuruniversity.in,

[2] Department of Mathematics, Jadavpur University, Kolkata-700032, INDIA
skuiry.math.rs@jadavpuruniversity.in, alakadas2012@gmail.com

**Abstract:** Object detection in natural scenes can be a challenging task. In many real-life situations, the visible spectrum is not suitable for traditional computer vision tasks. Moving outside the visible spectrum range, such as the thermal spectrum or the near-infrared (NIR) images, is much more beneficial in low visibility conditions, NIR images are very helpful for understanding the object's material quality. In this work, we have taken images with both the Thermal and NIR spectrum for the object detection task. As multi-spectral data with both Thermal and NIR is not available for the detection task, we needed to collect data ourselves. Data collection is a time-consuming process, and we faced many obstacles that we had to overcome. We train the YOLO v3 network from scratch to detect an object from multi-spectral images. Also, to avoid overfitting, we have done data augmentation and tune hyperparameters.

## 1    Introduction

In the last couple of decades, the field of computer vision has grown considerably. In terms of efficiency, pace, and scalability, many complex tasks such as object detection, localization, segmentation, and natural scene understanding have received a significant boost, particularly after implementing deep learning approaches such as convolutional neural networks [1]. However, deep learning approaches rely heavily on the availability of abundant quantities of data of good quality. Although visible light provides information close to what the human eye processes, it is unable to provide useful information in some cases. In a foggy atmosphere or a setting where the lighting is low, RGB systems can not provide adequate details, so there is a need for alternate imaging systems if such conditions have a fair chance of occurring. Thermal infrared imaging is also one such reliable device because since it captures heat signatures, infrared information can be independent of the efficiency of the light source(s). In warm environments, however, thermal infrared imaging can suffer and is based on costly camera hardware. Near-infrared cameras have been found to perform efficiently in situations that can not be used by RGB cameras due to lighting conditions, such as foggy environments. For nighttime conditions, near-infrared (nonvisible) illumination may also be introduced. Therefore, if we can integrate the additional information from Near-infrared and Thermal spectra into different deep learning tasks along with our existing visible spectra, it is expected to work better as the model gets more information on any object than the only visible spectrum. There are many [2] datasets available for classification purposes, but there are very few in the Multi-spectral domain. In Table 1, some of the well-known multi-spectral datasets have been mentioned. However, on these three spectrums, visible, near-infrared, and thermal, there are no datasets available publicly for the object recognition task.

So in this work, we have collected some Multi-spectral (RGB, NIR, and Thermal) natural scenes. The collected dataset not only brings into the learning process three different spectrums, but also other similar variables that work with different sensors, such as different focal lengths, viewpoints, noise sensitivity, resolution of images, sharpness, and so on. The main objectives of this paper are

1. Prepare a dataset of multi-spectral images, which contains RGB, Thermal, and NIR spectrum, of the same scene with annotation of 10 classes in YOLO and Pascal VOC format. The classes are as follows: Car, Motorcycle, Bicycle, Human, Building, Bush, Tree, Sign-board, Road, Window.
2. Prepare an augmented dataset of the dataset mentioned above, which followed the above format and classes.
3. Study of various object detection models where we have mainly focused on three YOLO models.
4. Using the YOLO v3 network, we have trained our dataset in three ways as follows:
    a. Thermal
    b. RGB and Thermal
    c. RGB, NIR, and Thermal
5. Discuss the performances of those models on the mentioned dataset and analyze the overall performances.

Some significant researches, [3][4][5][6] has been performed for object detection purposes on Multichannel Convolutional Neural Networks over the past decade. A related philosophy is still used in Multi-spectral imaging; however, the Multi-spectral imaging framework is very diverse. For instance, multi-spectral imaging in medicine [7] helps diagnose retinopathy and macular oedema until it affects the retina. It is used in surveillance for broad scanning areas [8] as well as for facial recognition because face recognition algorithms on Multi-spectral images [9] are found to work better. Multi-spectral data in the agriculture sector helps detect the variety [10] and chemical composition [11] of different plants which are useful for plant health monitoring, nutrient [12], water status, and pesticide requirements [13]. Multi-spectral data is also useful in detecting defects in the food processing industry[14], hazardous compounds in the chemical industry[15], etc.

The rest of the paper is organized as follows. In section "Experiment setup" we have briefly explained the data set collection and annotation process. In the next section, we have discussed the result and analysis performed on the YOLO v3 model performed on the collected data. The conclusion and future work has been discussed in the last section.

**Table 1**: Some of the Multi-spectral images datasets

| Authors | Year | Type | Spectrum |
| --- | --- | --- | --- |
| Takumi at al.[16] | 2017 | Street Images | RGB,NIR,MIR, FIR |
| Ha et al.[17] | 2017 | Street Images | RGB, Thermal |
| Aguilera et al.[18] | 2018 | Natural Images | RGB, NIR |
| Alldieck et al.[19] | 2016 | Street Images | RGB, Thermal |
| Brown et al.[20] | 2011 | Natural Images | RGB, NIR |
| Choe et al.[21] | 2018 | Natural Images | RGB, NIR |
| Davis et al.[22] | 2007 | Natural Images | RGB, Thermal |
| Hwang et al.[23] | 2015 | Natural Images | RGB, Thermal |
| Li et al.[24] | 2018 | Natural Images | RGB, Thermal |

## 2 Dataset Preparation

### 2.1 Dataset Collection

Image data within the specific wavelength ranges, captured using instruments which are sensitive to a particular wavelength or with the help of some filters by which the wavelengths may be parted, is called Multispectral Images. Our multi-spectral image dataset includes RGB, NIR (750–900 nm) and Thermal (10410-12510 nm) spectrum images. The difference in reflectivity of some particular objects, combined with decreased distortion and atmospheric haze in the NIR wavelength, helps to get details of the objects and the visibility is often improved [32]. In Thermal images, more heated objects are visualized well against less heated objects in the scene regardless of day-night or haze [33]. The details of the capturing devices are given below :

- Visible Spectrum : Nikkon D3200 DSLR Camera with Nikkon AF-S 3.5-5.6 G standard lens
- Near Infrared Spectrum : Watec WAT-902H2 Camera, 24mm lens ( SV-EGG-BOXH1X), Schnieder 093 IR Pass Filter (830nm)
- Thermal Spectrum: FLIR A655SC Thermal Camera

The raw image dataset containing multi-spectral images for the same scene is not readily available. No standard dataset is available freely related to the present work. There exist several available datasets either containing RGB, and NIR images [44] or RGB and Thermal images [43] for the same scene and also their annotation format is not uniform. As the required dataset is not available, we are forced to collect our dataset. The most challenging part of our work is preparing the same scene's dataset having natural images in three spectrums. We have collected 1060 images for each spectrum in the present work and totalling 3180 images of the 1060 scene.

## 2.2 Data Splitting

We have three spectral image datasets which we had collected on our own for the development of multi-spectral image data sets which are not readily available. We have used this dataset for object detection of 10 classes. The classes are Car, Motorcycle, Bicycle, Human, Building, Bush, Tree, Sign-board, Road, Window. The whole dataset is divided for each spectrum into six-part. One part out of six-part is completely kept separated for testing purposes so that we can understand the performance of our model. Left five parts are divided into 8:1 ratio for training and validation purposes. To increase the amount of data, we use the data augmentation technique, and those data are used only for training purposes. The test set consists of one complete scene which is not included in the training or validation set and 10% data of the training set.

## 2.3 Augmentation

As the collection of data is highly time-consuming and we need a significantly large amount of data to feed the deep neural network (DCNN) [36-39], we have taken the help of the data augmentation technique[45]. These are the most commonly used method for increasing the volume of the dataset significantly. It helps to reduce overfitting, to bring diversity in data, and to regularize the model. We have used Flip, Rotation, Scale, Crop, Translation, and Gaussian noise-based data augmentation techniques to randomly increase the volume of the dataset. Those augmentation techniques are standard and help to bring diversity in data substantially.

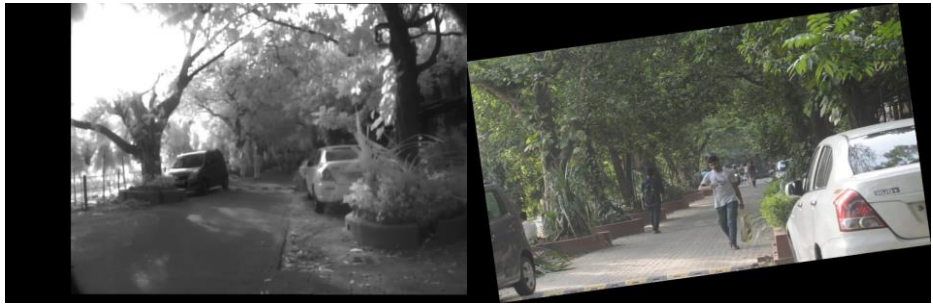

**Fig. 1.** Randomly translated NIR image and randomly rotated RGB image from our augmented dataset

5### 2.4 Annotation

The most important thing to be done before feeding the data to a deep neural network (DCNN) is annotation. Among the various annotation methods, we have used the bounding box technique. For completing the annotation, we have taken the help of labelling tools [29]. It is an open-source, free tool used for annotation in the bounding box method. As we have to feed the dataset to the YOLO v3 model, we have saved the annotations in darknet format.

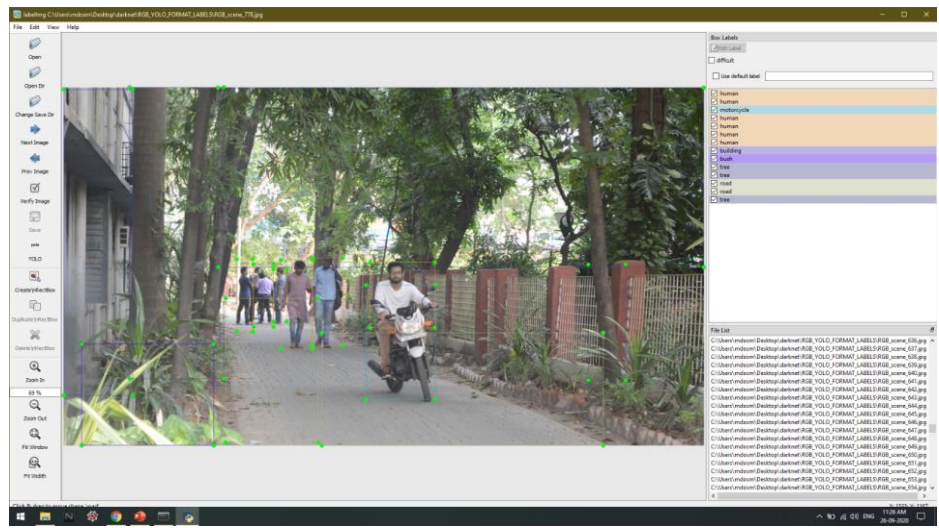

**Fig. 2.** Annotation in bounding box method using LabelImg tools

## 3 Result and Analysis

The natural scene is clicked randomly by three cameras simultaneously for the same scene multi-spectral dataset. All the images are pre-processed and taken together. Annotation is done for all the images, and then we split the dataset into train and test set. After augmentation, all the regular and augmented images and annotation of train part are being fed to the YOLO v3 network for training. The test portion images which are unknown to the model is used for detection. The whole process is presented in Figure 3.

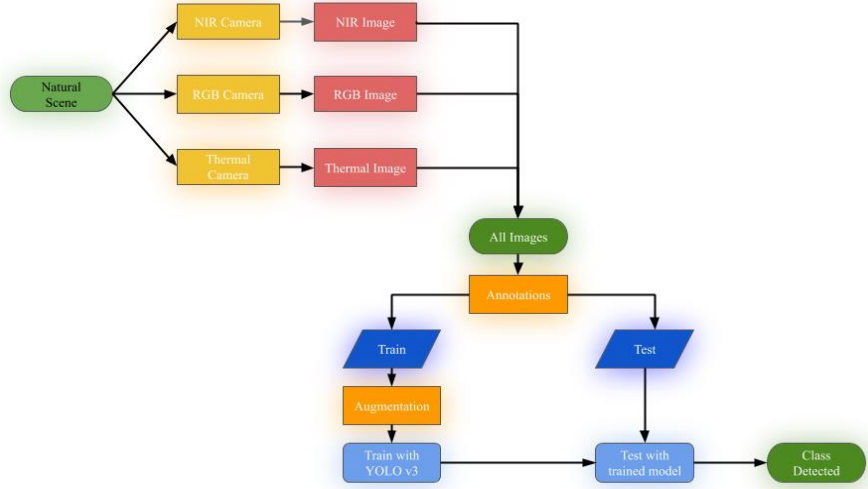

**Fig 3:** The proposed work

### 3.1 Experiment Setup

Among different existing object localization technique, YOLO and its variations are used heavily in the present days. YOLO v1 is a regression-based algorithm which uses the Darknet framework. YOLO v1 has several limitations as a result of the utilization of the YOLO v1 is restricted. It is unable to identify the tiny objects if they have appeared as a cluster. In the interest of improvement of the YOLO v1 network and overcome the limitations of YOLO v1, YOLO v2 [26] is developed. That incremental improvement of YOLO v2 or YOLO 9000 is termed as YOLO v3 [27]. The importance of an object detection algorithm is weighted by how accurate and quickly it can detect objects. YOLO v3 has done a great job and left every object detection algorithm behind in terms of speed and accuracy. It has everything to use for real-time object detection [34-35]. We have selected the YOLO v3 model architecture for our training purpose. The training was done on a LINUX (Ubuntu 18.04 LTS ) Operating system having 4 GB NVIDIA GPU ( GeForce GTX 1050 Ti) which helped us to train faster. Also, all the codes have been written and compiled in PyTorch environment. To evaluate the performance of the model, we have used the AP(Average Precision) and mAP (Mean Average Precision) [40] as a metric with IoU = 0.5. As an optimizer, we have used Adams Optimizer and fixed the learning rate into $7*10-4$. Firstly, we have trained only the THERMAL dataset, then THERMAL and RGB dataset together and finally we have trained THERMAL, RGB, and NIR together. For each training, we train our network with 380-400 epochs.

7### 3.2   THERMAL image Dataset Training

To measure the performance of the model, in Thermal dataset, we have kept one scene entirely out of training and validation purpose, which is used for testing the performance of the model. For testing, we have used 212 images. On the Thermal dataset, we get the $mAP_{IoU=0.5} = 53.4\%$, which is comparable to YOLO v3 results on COCO [42] dataset. The results are given in the tabular form, and examples of detection are given below:

**Table 2.** AP is calculated at the IoU threshold of 0.5 of all classes on the Thermal dataset and mAP@0.5 are given in the table.

| Class | $AP_{50}$ |
|---|---|
| Car | 67.1 |
| Motorcycle | 43.5 |
| Bicycle | 5.5 |
| Human | 21.4 |
| Building | 61.5 |
| Bush | 62.6 |
| Tree | 79.3 |
| Sign-board | 27.5 |
| Road | 73.4 |
| Window | 92.5 |
| **$mAP_{50}$** | **53.4** |

If we closely observe the above table (Table 2), which indicates the result obtained from the test set of Thermal images, we will find that the $AP_{50}$ value of bicycle class too low because the number of samples of that particular class was significantly less. Other classes are giving a moderate $AP_{50}$ value, whereas the window class is giving the highest $AP_{50}$ value. The result depends on the scene's temperature, as the camera automatically normalizes the overall scene temperature if it finds a temperature a considerable variance of temperature (like some burning object, sky scene, etc.). The overall $mAP_{50}$ is quite satisfactory, and it is comparable to the YOLO v3 mAP for the COCO dataset. The detection of the object on test samples are given below:

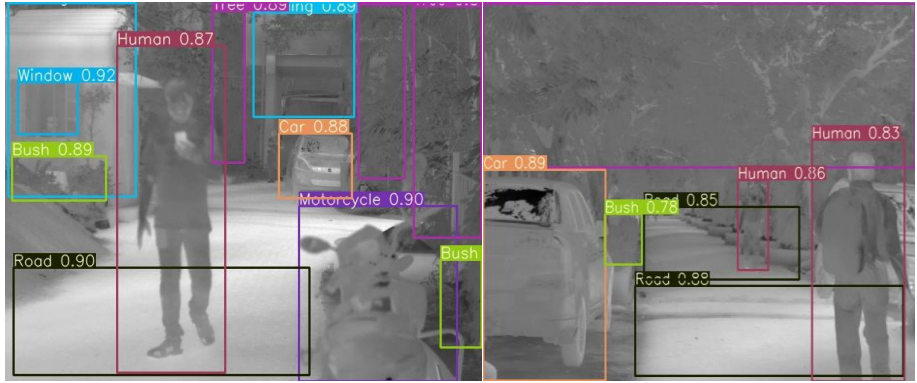

**Fig. 4.** Object detected on Thermal Images from test set. Non-maximal Suppression = 0.5, Confidence Threshold = 0.75, IoU = 0.5

### 3.3 RGB and THERMAL Image Dataset Training

As discussed, to measure the model performance of RGB and Thermal dataset, we have kept one scene completely out of training and validation purpose, which is used for testing the performance of the model. For testing, we have used 240 images. No training set data is used for testing. On Thermal dataset we get the $mAP_{IoU=0.5} = 46.4\%$. The results are given in the tabular form, and examples of detection are given below:

**Table 3.** AP is calculated at the IoU threshold of 0.5 of all classes on RGB, and the Thermal dataset and mAP@0.5 are given in the table.

| Class | $AP_{50}$ |
|---|---|
| Car | 69.9 |
| Motorcycle | 52.6 |
| Bicycle | 59.4 |
| Human | 78.3 |
| Building | 23.4 |
| Bush | 35.4 |
| Tree | 48.4 |
| Sign-board | 14.5 |
| Road | 51.4 |
| Window | 30.5 |
| **$mAP_{50}$** | **46.4** |

Adding RGB images with Thermal images for training shows a significant change in the $AP_{50}$ of every class. Here a drastic change is noticeable in $AP_{50}$ for the bicycle class, because of increased visibility due to the visual spectrum. Although AP of some classes, like sign-board, got reduced by a significant margin. This thing happened due to the imbalance in the number of classes. The overall $mAP_{50}$ = 46.4% is a good achievement for multi-spectral classes. The examples of detection of objects on RGB and Thermal images using this model is presented below:

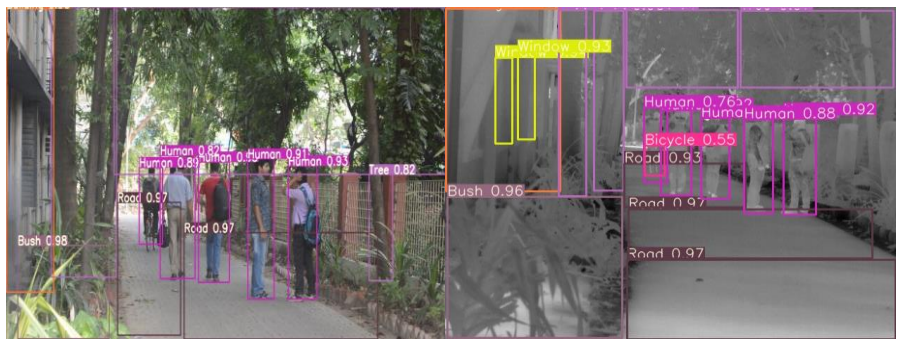

**Fig. 5.** Object detected on RGB and Thermal Images from test set. Non-maximal Suppression = 0.5, Confidence Threshold = 0.4, IoU = 0.5

### 3.4 THERMAL, NIR and RGB Dataset Training

We tried to observe and work on improving the performance of the YOLO v3 model on the Multi-spectral dataset. To measure the model performance of the Multi-spectral dataset, we have kept one scene completely out of training and validation purpose, which is used for testing the performance of the model. For testing, we have used 170 images. On Thermal dataset we get the $mAP_{IoU=0.5}$ = 43.4%. The results are given in the tabular form, and examples of detection are given below:

**Table 4.** AP is calculated at the IoU threshold of 0.5 of all classes on Thermal, NIR, and RGB dataset, and mAP@0.5 are given in the table.

| Class | $AP_{50}$ |
| --- | --- |
| Car | 30.9 |
| Motorcycle | 55.1 |
| Bicycle | 61.5 |
| Human | 45.8 |
| Building | 39.4 |
| Bush | 40.7 |
| Tree | 31.5 |
| Sign-board | 47.1 |
| Road | 50.0 |
| Window | 31.8 |
| **$mAP_{50}$** | **43.4** |

After training the model with RGB, NIR, and Thermal images, the performance is slightly reduced. As we started adding more spectral images, the overall performance started to reduce, but each class's AP became stable. The AP50 of all classes is moderate, and no class has got significantly reduced AP50. The class car has got the least AP50 =30.9%. The overall mAP50 = 43.4%. Using our model, the object detection result is shown below:

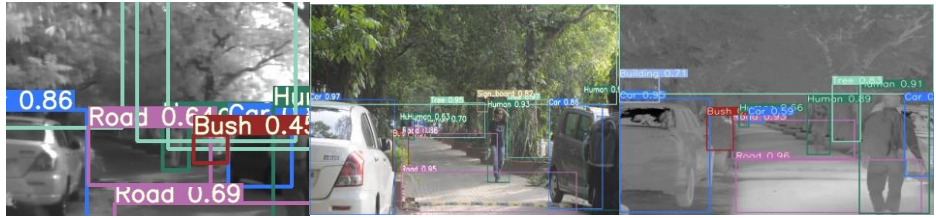

**Fig. 6.** Object detected on NIR, RGB and THERMAL Images respectively from test set. Non-maximal Suppression = 0.5, Confidence Threshold = 0.4, IoU = 0.5

Although the training of only the Thermal image dataset gives a quite good output on the test set the output of the multi-spectral dataset is not that bad. As we started adding different spectral images, the performance started degrading.

**Table 5.** Comparison of performance on the different spectral dataset

| Input Spectrum | mAP$_{50}$ |
| --- | --- |
| Thermal | 53.4 |
| RGB and Thermal | 46.4 |
| RGB, NIR, and Thermal | 43.4 |

Mean Average Precision values of only Thermal images are reasonably good. For only Thermal images, mAP50 = 53.4%. The mAP of RGB and Thermal images is 46.4%. When we add one more spectrum (i.e., NIR), the mAP is reduced to 43.4%. The accuracy of the models lowers, as we append more spectral image datasets. The performance is reduced due to less visibility and less differentiable pixel values of Thermal and NIR images. More accuracy can be obtained by increasing the volume of the dataset by a large margin.

First of all, we know that thermography depends on the heat emitting from the object. The object emits more heat is more clearly visible. In the case of an RGB image, the object is detected depending on the pattern and the structure. Moreover, for the NIR image, it depends on the reflectivity of certain objects. For some objects, the AP increase but, in some cases, it decreases. Low AP for Bicycle class for only thermal image signifies that the lower temperature of the class Bicycle as the data collected in the winter season, and it started increasing when we add RGB and NIR spectrum respectively. The AP of some objects like Road decreases when we add more spectrum because the model is extracting more complex features from different spectrums. For some objects like Sign-board, it is shown that after adding RGB, it decreases significantly while adding the NIR spectrum it increases. This is because of the class imbalance in our dataset, which will be adjusted before publishing the dataset publicly.

## 4 Conclusion and Future Work

Firstly, we have collected data on multi-spectral natural images. The spectral bands included Thermal, NIR, and RGB. After collecting and pre-processing the data, we annotate them. To avoid overfitting and increase the amount of data, we perform data augmentation. When the dataset is ready to feed into the network, we prepare the YOLO v3 model to perform training. After training multiple times and tuning hyperparameters, we conclude with the aforementioned results and detection examples.

Our model performance gives a brief outline that the multi-spectrum dataset can be used for object detection. Although the model's performance is not so outstanding, it can be improved by performing better hyperparameter tuning and adding more data into the dataset. The main problems of this dataset are class imbalance and less amount of data. As the dataset has a significantly less amount of data, we had to deal with the model's overfitting. As the instances per class fluctuate very sharply, the model unable to give optimal results. As the number of spectrum increases, the model struggles to perform well. This happens due to a lack of data and class imbalance. In the COCO dataset, we have seen a considerable amount of data with a very large number of instances for a single class, which helps the model perform much more accurately.

In the future, we can increase the amount of data that can handle the class imbalance. We can merge three spectrum images in a single image called image registration. Here we can feed a network with five layers (RGB = 3, NIR = 1, Thermal = 1, totalling = 5) of data at a time, which can help us to improve accuracy. Not only object detection, but we can also use the merged data for object material detection, object temperature detection as well. The model performance can be improved significantly by introducing an ensembling among the different state of the art models[46]. After some processing and adding more images into the dataset, the whole dataset will be made public for research purposes.

## 5 Acknowledgement

We want to thank the entire team of people that made the collection of this dataset possible.
 – Image Capturing:  Priyam Sarkar
 – Data Preprocessing: Priyam Sarkar
 – Data Annotation: Shubhadeep Bhowmick

This work is supported by the project sponsored by SERB (Government of India, order no. SB/S3/EECE/054/2016) (dated 25/11/2016), and carried out at the Centre for Microprocessor Application for Training Education and Research, CSE Department, Jadavpur University. The second author would like to thank The Department of Science and Technology for their INSPIRE Fellowship program (IF170641) for financial support.